# Solving Asymmetric Decision Problems with Influence Diagrams

**Runping Qi**
Department of Computer Science
UBC
Vancouver B. C. Canada V6T 1Z4
E-mail: qi@cs.ubc.ca

**(Nevin) Lianwen Zhang**
Department of Computer Science
HUST
Hongkong
E-mail: lzhang@cs.ust.hk

**David Poole***
Department of Computer Science
UBC
Vancouver B. C. Canada V6T 1Z4
E-mail: qi@cs.ubc.ca

## Abstract

While influence diagrams have many advantages as a representation framework for Bayesian decision problems, they have a serious drawback in handling asymmetric decision problems. To be represented in an influence diagram, an asymmetric decision problem must be symmetrized. A considerable amount of unnecessary computation may be involved when a symmetrized influence diagram is evaluated by conventional algorithms. In this paper we present an approach for avoiding such unnecessary computation in influence diagram evaluation.

## 1 INTRODUCTION

Decision trees were used as a simple tool both for problem modeling and optimal policy computation in the early days of decision analysis (Raiffa 1968). A decision tree explicitly depicts all scenarios of the problem and specifies the "utility" the agent can get in each scenario. An optimal policy for a decision problem can be computed from the decision tree representation of the problem by a simple "average-out-and-fold-back" method.

Though conceptually simple, decision trees have a number of drawbacks. First, the dependency/independency relationships among the variables in a decision problem cannot be represented in a decision tree. Second, a decision tree specifies a particular order for the assessment on the probability distributions of the random variables in the decision problem. This order is in most cases not a natural assessment order. Third, the size of a decision tree for a decision problem is exponential in the number of variables of the decision problem. Finally, a decision tree is not easily adaptable to changes in a decision problem. If a slight change is made in a problem, one may have to draw a decision tree anew.

Influence diagrams were proposed as an alternative to decision trees for decision analysis (Howard and Matheson, 1984, Miller et.al. 1976). As a representation framework, influence diagrams do not have the aforementioned drawbacks of decision trees. The influence diagram representation is expressive enough to explicitly describe the dependency/independency relationships among the variables in the decision problem; it allows a more natural assessment order on the probabilities of the uncertain variables; it is compact; and it is easy to adapt to the changes in the problem.

However, in comparison with decision trees, influence diagrams have one disadvantage in representing asymmetric decision problems (Covaliu and Oliver 1992, Fung and Shachter 1990, Phillips 1990, Shachter 1986, Smith et al. 1993). Decision problems are usually asymmetric in the sense that the set of possible outcomes of a random variable may vary depending on different conditioning states, and the set of legitimate alternatives of a decision variable may vary depending on different information states. To be represented as an influence diagram, an asymmetric decision problem must be "symmetrized" by adding artificial states and assuming degenerate probability distributions (Smith et al. 1993). This symmetrization results in two problems. First, the number of information states of decision variables are increased. Among the information states of a decision variable, many are "impossible" (having zero probability). The optimal choices for these states need not be computed at all. However, they are computed by conventional influence diagram evaluation algorithms (Shachter 1986, Smith et al. 1993, Shachter and Peot 1992, Zhang and Poole 1992, Zhang et al. 1993). Second, for each information state of a decision variable, because the legitimate alternatives may constitute only a subset of the frame of the decision variable, an optimal choice is chosen from only a subset of the frame, instead of the entire frame. However, conventional influence diagram algorithms have to consider all alternative in order to compute an optimal choice for a decision in any of its information states. Thus, it is evident that conventional influence diagram evaluation algorithms involve unnecessary computation.

In this paper, we present an approach for overcoming the aforementioned disadvantage of influence diagrams. Our approach consists of two independent components: a simple extension to influence diagrams and a top–down method for influence diagram evaluation. Our extension allows explicitly expressing the fact that some decision variables have different frames in different information states. Our method, similar to Howard and Matheson's (1984), evaluates an influence diagram in two conceptual steps: it first maps an influence diagram into a decision tree (Qi 1994) in such a way that an optimal solution tree of the decision tree corresponds to an optimal policy of the influence diagram. Thus the problem of computing an optimal policy is reduced to the problem of searching for an optimal solution tree of a decision tree, which can be accomplished by various algorithms (Qi 1994). Like Howard and Matheson's method, ours avoids computing optimal choices for decision variables in impossible states. Furthermore, our method has two advantages over Howard and Matheson's. First, the size of the intermediate decision tree generated by our method is much smaller than that generated by Howard and Matheson's for the same influence diagram. Second, our method provides a clean interface between influence diagram evaluation and Bayesian net evaluation so that various well-established algorithms for Bayesian net evaluation can be used in influence diagram evaluation. This method works for influence diagrams with or without our extension.

The rest of this paper is organized as follows. The next section introduces influence diagrams. Section 3 uses an example to illustrate the disadvantage that influence diagrams and their solution algorithms have with asymmetric decision problems. In Section 4, we present our approach for overcoming the disadvantage. Section 5 gives an analysis on how much can be saved by exploiting asymmetry in decision problems. Section 6 discusses related work and Section 7 concludes the paper.

## 2  INFLUENCE DIAGRAMS

The following definition for influence diagrams is borrowed from (Zhang et al. 1993). An influence diagram $\mathcal{I}$ is defined as a quadruple $\mathcal{I} = (X, A, \mathcal{P}, \mathcal{U})$ where

- $(X, A)$ is a directed acyclic graph with node set $X$ and arc set $A$. The node set $X$ is partitioned into random node set $C$, decision node set $D$ and value node set $U$. All the nodes in $U$ have no child.

  Each decision node or random node has a set, called the *frame*, associated with it. The frame of a node consists of all the possible outcomes of the (decision or random) variable denoted by the node. For any node $x \in X$, we use $\pi(x)$ to denote the parent set of node $x$ in the graph and use $\Omega_x$ to denote the frame of node $x$. For any subset $J \subseteq C \cup D$, we use $\Omega_J$ to denote the Cartesian product $\prod_{x \in J} \Omega_x$.

- $\mathcal{P}$ is a set of probability distributions $P\{c|\pi(c)\}$ for all $c \in C$. For each $o \in \Omega_c$ and $s \in \Omega_{\pi(c)}$, the distribution specifies the conditional probability of event $c = o$, given that[1] $\pi(c) = s$.

- $\mathcal{U}$ is a set $\{g_v : \Omega_{\pi(v)} \to \mathcal{R} | v \in U\}$ of *value functions* for the value nodes, where $\mathcal{R}$ is the set of the real.

For a decision node $d_i$, a mapping $\delta_i : \Omega_{\pi_{d_i}} \to \Omega_{d_i}$ is called a *decision function* for $d_i$. The set of all the decision functions for $d_i$, denoted by $\Delta_i$, is called the *decision function space* for $d_i$. Let $D = \{d_1, ..., d_n\}$ be the set of decision nodes in influence diagram $\mathcal{I}$. The Cartesian product $\Delta = \prod_{i=1}^{n} \Delta_i$ is called the *policy space* of $\mathcal{I}$.

For a decision node $d_i$, a value $x \in \Omega_{\pi(d_i)}$ is called an information state of $d_i$, and a mapping $\delta_i : \Omega_{\pi(d_i)} \to \Omega_{d_i}$ is called a *decision function* for $d_i$. The set of all the decision functions for $d_i$, denoted by $\Delta_i$, is called the *decision function space* for $d_i$. The Cartesian product of the decision function spaces for all the decision nodes is called the *policy space* of $\mathcal{I}$. We denote it by $\Delta$.

Given a policy $\delta = (\delta_1, \ldots, \delta_k) \in \Delta$ for $\mathcal{I}$, a probability $P_\delta$ can be defined over the random nodes and the decision nodes as follows:

$$P_\delta(C, D) = \prod_{c \in C} P(c|\pi(c)) \prod_{i=1}^{k} P_{\delta_i}(d_i|\pi(d_i)), \quad (1)$$

where $P(c|\pi(c))$ is given in the specification of the influence diagram, while $P_{\delta_i}(d_i|\pi(d_i))$ is given by $\delta_i$ as follows:

$$P_{\delta_i}(d_i|\pi(d_i)) = \begin{cases} 1 & \text{when } \delta_i(\pi(d_i)) = d_i, \\ 0 & \text{otherwise} \end{cases} \quad (2)$$

For any value node $v$, $\pi(v)$ must consist of only decision and random nodes, since value nodes do not have children. Hence, we can talk about $P_\delta(\pi(v))$. The *expectation of the value node $v$ under $P_\delta$*, denoted by $E_\delta[v]$, is defined as follows:

$$E_\delta[v] = \sum_{\pi(v)} P_\delta(\pi(v)) f_v(\pi(v)).$$

The summation $E_\delta = \sum_{v \in U} E_\delta[v]$ is called the *value* of $\mathcal{I}$ under the policy $\delta$. The maximum of $E_\delta$ over all the possible policies $\delta$ is the *optimal expected value* of $\mathcal{I}$. An *optimal policy* is a policy that achieves the optimal expected value. To *evaluate* an influence diagram is to determine its optimal expected value and to find an optimal policy.

---

[1] In this paper, for any variable set $J$ and any element $e \in \Omega_J$, we use $J = e$ to denote the set of assignments that assign an element of $e$ to the corresponding variable in $J$.

An influence is *regular* if there exists a total ordering among all the decision nodes. The results presented in this paper are applicable to regular *stepwise decomposable influence diagrams* (Qi 1993, Qi and Poole 1993, Zhang and Poole 1992). We shall, however, limit the exposition only to regular influence diagrams with a single value node for simplicity.

# 3 WHY INFLUENCE DIAGRAMS ARE NOT GOOD FOR ASYMMETRIC DECISION PROBLEMS

In this section, we illustrate by an example the disadvantages of conventional influence diagrams with asymmetric decision problems. We use the used car buyer problem (Howard 1962) because it is a typical asymmetric decision problem and it has been used by other researchers (Shenoy 1993, Smith *et al.* 1993).

## 3.1 THE USED CAR BUYER PROBLEM

Joe is considering to buy a used car. The marked price is $1000, while a three years old car of this model worths $1100, if it has no defect. Joe is uncertain whether the car is a "peach" or a "lemon". But Joe knows that, of the ten major subsystems in the car, a peach has a defect in only one subsystem whereas a lemon has a defect in six subsystems. Joe also knows that the probability for the used car being a peach is 0.8 and the probability for the car being a lemon is 0.2. Finally, Joe knows that it will cost him $40 to repair one defect and $200 to repair six defects.

Observing Joe's concern about the possibility that the car may be a lemon, the dealer offers an "anti-lemon guarantee" option. For an additional $60, the anti-lemon guarantee will cover the full repair cost if the car is a lemon, and cover half of the repair cost otherwise. At the same time, a mechanic suggests that some mechanical examination should help Joe determine the car's condition. In particular, the mechanic gives Joe three alternatives: test the steering subsystem alone at a cost of $9; test the fuel and electrical subsystems at a total cost of $13; a two-test sequence in which, the transmission subsystem will be tested at a cost of $10, and after knowing the test result, Joe can decide whether to test the differential subsystem at an additional cost of $4. All tests are guaranteed to detect a defect if one exists in the subsystem(s) being tested.

## 3.2 INFLUENCE DIAGRAM REPRESENTATION FOR THE USED CAR BUYER PROBLEM

An influence diagram for the used car problem is shown in Fig. 1. The random variable CC represents the car's condition. The frame for CC has two elements: peach and lemon. The variable has no parent in the graph, thus, we specify its prior probability distribution in Table 1.

The decision variable $T_1$ represents the first test decision. The frame for $T_1$ has four elements: nt, st, f&e and tr, representing respectively the options of performing no test, testing the steering subsystem alone, testing the fuel and electrical subsystems, and testing the transmission subsystem with a possibility of testing the differential subsystem next.

The random variable $R_1$ represents the first test results. The frame for $R_1$ has four elements: nr, zero, one and two representing respectively the four possible outcomes of the first test: no result, no defect, one defect and two defects. The probability distribution of the variables, conditioned on $T_1$ and CC, is given in Table 2.

The decision variable $T_2$ represents the second test decision. The frame for $T_2$ has two elements: nt and diff, denoting the two options of performing no test and testing the differential subsystem.

The random variable $R_2$ represents the second test results. The frame for the random variable $R_2$ has three elements: nr, zero and one, representing respectively the three possible outcomes of the second test: no result, no defect and one defect. The probability distribution of the variables, conditioned on $T_1$, $R_1$, $T_2$ and CC, is given in Table 3.

The decision variable B represents the purchase decision. The frame for B has three elements: b̄, b and g, denoting respectively the options of not buying the car, buying the car without the anti-lemon guarantee and buying the car with the anti-lemon guarantee.

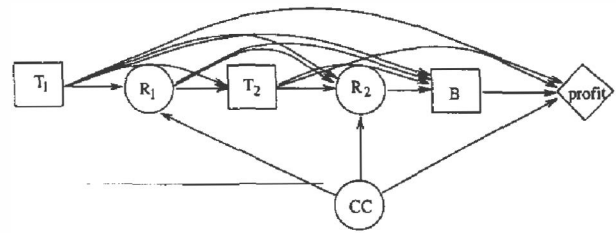

Figure 1: An influence diagram for the used car buyer problem

The used car buyer problem is asymmetric in a number of aspects. First, the set of the possible outcomes of the first test result varies, depending on the choice for the first test. If the choice for the first test is nt, then there is only one possible outcome for the first test result — **nr** (representing no result). If the choice for the first test is **st** or **tr**, then there are two possible outcomes for the first test result — **zero** and **one** (representing no defect and one defect, respectively). If the choice for the first test is **f&e**, then there are three possible outcomes for the first test result — **zero**, **one** and **two** (representing no defect, one defect and two defects, respectively). However, in the influence diagram representation, the frame of the variable $\mathtt{R_1}$ is a common set of outcomes for all the three cases. The impossible combinations of the test choices and the test results are characterized by assigning zero probability to them (as shown in Table 2). A similar discussion is applicable to the variable $\mathtt{R_2}$. Second, from the problem statement we know that testing differential subsystem is possible only in the states where the first test performed is on the transmission subsystem. However, in the influence diagram representation, it appears that the second test is possible in any situation, while the fact that the option of testing differential subsystem is not available in some situations is characterized by assigning unit probability to outcome **nr** of the variable $\mathtt{R_2}$ conditioned on these situations. Third, when we examine the information states of the decision variable $\mathtt{T_2}$, we will see many combinations of test options and test results are impossible. For example, if Joe first tests the transmission subsystem, it is impossible to observe **nr** and **two**. If the influence diagram is evaluated by conventional algorithms, an optimal choice for the second test will be computed for each of the information states, including many impossible states. Similar argument is applicable to the decision variable **B**. Because it is not necessary to compute optimal choices of a decision variables for impossible states, it is desirable to avoid the computation.

Table 1: The prior probability distribution of the car's condition $P\{\mathtt{CC}\}$

| CC | prob |
|---|---|
| peach | 0.8 |
| lemon | 0.2 |

Table 2: The probability distribution of the first test result $P\{\mathtt{R_1}|\mathtt{T_1},\mathtt{CC}\}$

| $\mathtt{T_1}$ | CC | $\mathtt{R_1}$ | prob |
|---|---|---|---|
| nt | – | nr | 1.0 |
| nt | – | others | 0 |
| st | – | nr | 0 |
| st | – | two | 0 |
| st | peach | zero | 0.9 |
| st | peach | one | 0.1 |
| st | lemon | zero | 0.4 |
| st | lemon | one | 0.6 |
| f&e | – | nr | 0 |
| f&e | peach | zero | 0.8 |
| f&e | peach | one | 0.2 |
| f&e | peach | two | 0 |
| f&e | lemon | zero | 0.13 |
| f&e | lemon | one | 0.53 |
| f&e | lemon | two | 0.33 |

Table 3: The probability distribution of the second test result $P\{\mathtt{R_2}|\mathtt{T_1},\mathtt{R_1},\mathtt{T_2},\mathtt{CC}\}$

| $\mathtt{T_1}$ | $\mathtt{R_1}$ | $\mathtt{T_2}$ | CC | $\mathtt{R_2}$ | prob |
|---|---|---|---|---|---|
| nt | – | – | – | nr | 1.0 |
| nt | – | – | – | others | 0 |
| st | – | – | – | nr | 1.0 |
| st | – | – | – | others | 0 |
| f&e | – | – | – | nr | 1.0 |
| f&en | – | – | – | others | 0 |
| tr | nr | – | – | nr | 1.0 |
| tr | nr | – | – | others | 0 |
| tr | two | – | – | nr | 1.0 |
| tr | two | – | – | others | 0 |
| tr | – | nt | – | nr | 1.0 |
| tr | – | nt | – | others | 0 |
| tr | zero | diff | peach | zero | 0.89 |
| tr | zero | diff | peach | one | 0.11 |
| tr | zero | diff | lemon | zero | 0.67 |
| tr | zero | diff | lemon | one | 0.33 |
| tr | one | diff | peach | zero | 1.0 |
| tr | one | diff | peach | one | 0 |
| tr | one | diff | lemon | zero | 0.44 |
| tr | one | diff | lemon | one | 0.56 |

## 4   OUR SOLUTION

In this section, we present an approach for overcoming the aforementioned disadvantage of influence diagrams. Our approach consists of two independent components: a simple extension to influence diagrams and a top-down method for influence diagram evaluation.

Our extension allows explicitly expressing the fact that some decision variables have different frames in different information states. We achieve this by introducing a *framing function* for each decision variable, which characterizes the available alternatives for the decision variable in different information states. With the help of framing functions, our solution algorithm effectively ignores the unavailable alternatives when computing an optimal choice for a decision variable in any information state. Our extension is inspired by the concepts of *indicator valuations* and *effective frames* proposed by Shenoy (1993).

Conceptually, our evaluation method, similar to Howard and Matheson's method (Howard and Math-

eson 1984), consists of two steps: in order to evaluate an influence diagram, a decision tree is generated and the evaluation is then carried out on the decision tree. The first step will be described in this section. The second step can be carried out either by the simple "average-out-and-fold-back" method (Raiffa 1968), or by a top-down search algorithm (Qi 1994). An advantage of using a search algorithm is that the two steps of tree generation and optimal policy computation can be combined into one, and only a portion of the tree needs to be generated, due to heuristic search. Our method successfully avoids the unnecessary computations by pruning those impossible states and ignoring those unavailable alternatives for the decision variables.

In comparison with than Howard and Matheson's method, ours has two distinct advantages. First, for the same influence diagram, our method generates a much smaller decision tree. Second, our method provides a clean interface to utilizing efficient Bayesian net algorithms (Lauritzen and Spiegelhalter 1988, Pearl 1988).

### 4.1 EXTENDING INFLUENCE DIAGRAMS

We extend influence diagrams by introducing *framing functions* to the definition given in Section 2. With this extension, an influence diagram $\mathcal{I}$ is a tuple $\mathcal{I} = (X, A, \mathcal{P}, \mathcal{U}, \mathcal{F})$ where $X, A, \mathcal{P}, \mathcal{U}$ have the same meaning as before, and $\mathcal{F}$ is a set $\{f_d : \Omega_{\pi(d)} \to 2^{\Omega_d}\}$ of *framing functions* for the decision nodes.

The framing functions express the fact that the legitimate alternative set for a decision variable may vary in different information states. More specifically, for a decision variable $d$ and an information state $s \in \Omega_{\pi(d)}$, $f_d(s)$ is the set of the legitimate alternatives the decision maker can choose for $d$ in information state $s$. Following Shenoy (1993), we call $f_d(s)$ the *effective frame* of decision variable $d$ in information state $s$.

Similarly, we define a decision function for a decision node $d_i$ as a mapping $\delta_i : \Omega_{\pi(d_i)} \to \Omega_{d_i}$. In additional, $\delta_i$ must satisfy the following constraint: For each $s \in \Omega_{\pi(d_i)}$, $\delta_i(s) \in f_{d_i}(s)$. In words, the choice prescribed by a decision function for a decision variable $d$ in an information state must be a legitimate alternative.

In the used car problem, the framing functions for the first test decision and the purchase decision are simple — they map every information state to the corresponding full frames.

The frame function for the second test decision can be specified as follows:

$$f_{T_2}(X) = \begin{cases} \{\text{nt, diff}\} & \text{if } \sigma_{T_1}(X) = \text{tr} \\ \{\text{nt}\} & \text{otherwise.} \end{cases}$$

### 4.2 CONSTRUCTING DECISION TREES FROM INFLUENCE DIAGRAMS

In the decision tree generated by our method for an influence diagram, a choice node corresponds to an information state of a decision variable, and a chance node corresponds to an uncertain state resulting from choosing an alternative for a decision variable in an information state. Two states are *consistent* if the variables common to both states have the same outcomes.

The decision tree is recursively specified as follows:

- Initially, the root, a chance node representing the empty state, is in the decision tree.

- For each information state $S$ of the first decision variable $d_1$, there is a choice node, as a child of the root in the decision tree. The arc from the root to the node is labeled with the probability $P\{\pi(d_1) = S\}$. A choice node in the decision tree is pruned if the probability on the arc to it is zero.

- Let $N$ be a choice node not pruned in the decision tree, and $S_N$ be the information state associated with $N$. Assume that $S_N$ is for decision variable $d$. Then, $N$ has $|f_d(S_N)|$ children, each corresponding to an alternative in $f_d(S_N)$. These children are all leaf nodes if $d$ is the last decision variable. Otherwise, they are chance nodes. The node corresponding to alternative $a \in f_d(S_N)$ represents the state $\pi(d) = S_N, d = a$.

- Let $N$ be a chance node representing a state $\pi(d_{i-1}) = S_N, d_{i-1} = a$, and let $\mathcal{A}$ be the subset of the information states of decision variable $d_i$ which are consistent with $\pi(d_{i-1}) = S_N, d_{i-1} = a$. Node $N$ has $|\mathcal{A}|$ children, each being a choice node representing an information state in $\mathcal{A}$. Let $S$ be the information state represented by a child of $N$. The arc from $N$ to the child is labeled with the conditional probability $P\{\pi(d_i) = S | \pi(d_{i-1}) = S_N, d_{i-1} = a\}$.

In the above specification, we effectively prune all of the impossible information states for all decision variables and ignore the unavailable alternatives to decision variables.

We have not specified how to compute the probabilities on the arcs from chance nodes nor how to compute the values associated with the leaf nodes. As illustrated in (Qi and Poole 1993), various well established Bayesian Net algorithms can be employed for computing the probabilities, and computing the values associated with the leaf nodes, which normally involve only small portions of the influence diagram. In particular, in order to further exploit asymmetry, Smith's method (Smith *et al.* 1993) can also be used for computing those probabilities.

496  Qi, Zhang, and Poole

## 5 HOW WELL OUR ALGORITHM DOES FOR THE USED CAR BUYER PROBLEM

When applying our algorithm to the used car buyer problem, a decision tree shown in Fig. 2 is generated. In the graph, the leftmost box represents the only situation in which the first test decision is to be made. The boxes in the middle column correspond to the information states in which the second test decision is to be made. Similarly, the boxes in the right column correspond to the information states in which the purchase decision is to be made. From the figure we see that among those nodes corresponding to the information states of the second test, all but two have only one child because the effective frames of the second test in the corresponding information states have only a single element. Making use of the framing function this way is equivalent to six prunings, each cutting a subtree under a node corresponding to an information state of the second test. Those shadowed boxes correspond to the impossible states. Our algorithm effectively detects those impossible states and prune them when they are created. Each of such pruning amounts to cutting a subtree under the corresponding node. Consequently, our algorithm does not compute optimal choices for a decision node for those impossible states. For the used car buyer problem, our algorithm computes optimal choices for the purchase decision for only 12 information states, and optimal choices for the second test for only 8 information states (among which six can be computed trivially). These constitute the minimal information state set one has to consider in order to compute an optimal policy for the used car buyer problem. This suggests that, as far as decision making concerned, our method exploits asymmetry to the maximum extent. In contrast, whereas those algorithms that do not exploit asymmetry will compute the optimal choices for the purchase decision for 96 ($4 \times 4 \times 2 \times 3$) information states and will compute optimal choices for the second test for 16 information states.

## 6 RELATED WORK ON HANDLING ASYMMETRIC DECISION PROBLEMS

Recognizing that influence diagrams are not effective asymmetric decision problems, several researchers have recently proposed alternative representations.

Fung and Shachter (1990) propose *contingent influence diagrams* for explicitly expressing asymmetry of decision problems. In that representation, each variable is associated with a set of contingencies, and associated with one relation for each contingence. These relations collectively specify the conditional distribution of the variable.

Covaliu and Oliver (1992) propose a different represen-tation for representing decision problems. This representation uses a *decision diagram* and a *formulation table* to specify a decision problem. A decision diagram is a directed acyclic graph whose directed paths identify all possible sequences of decisions and events in a decision problem. In a sense, a decision diagram is a degenerate decision tree in which paths having a common sequence of events are collapsed into one path (Covaliu and Oliver 1992). Numerical data are stored in the formulation table.

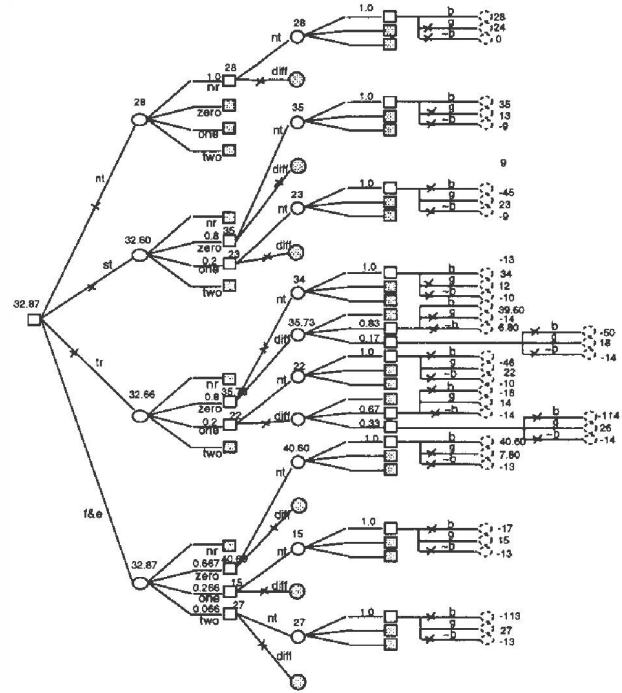

Figure 2: A decision tree generated for the used car buyer problem

Shenoy (1993) proposes a "factorization" approach for representing degenerate probability distributions. In that approach, a degenerate probability distribution over a set of variables is decomposed into several factors over subsets of the variables such that the their "product" is equivalent to the original distribution.

Smith *et al.* (1993) present some interesting progress towards exploiting asymmetry of decision problems. They observe that an asymmetric decision problem often has some degenerate probability distributions, and that the influence diagram evaluation can be sped up if these degenerate probability distributions are used properly. Their philosophy is analogous to the one behind various algorithms for sparse matrix computation. In their proposal, a conventional influence diagram is used to represent a decision problem at the level of relation. In addition, they propose to use a decision tree–like representation to describe the conditional probability distributions associated with the random variables in the influence diagram. The deci-sion tree-like representation is effective for economically representing degenerate conditional probability distributions. They propose a modified version of Shachter's algorithm (Shachter 1986) for influence diagram evaluation, and show how the decision tree-like representation can be used to increase the efficiency of *arc reversal*, a fundamental operation used in Shachter's algorithm. However, their algorithm cannot avoid computing optimal choices for decision variables with respect to impossible information states.

## 7 CONCLUSIONS

In this paper we analyzed a drawback of influence diagrams with asymmetric decision problems, which induces some unnecessary computation in solving asymmetric decision problems through influence diagram evaluation. We presented an approach for overcoming the drawback. Our approach consists of a simple extension to influence diagrams and a top-down method for influence diagram evaluation. The extension facilitates expressing asymmetry in influence diagrams. The top-down method effectively avoids unnecessary computation.

### Acknowledgement

The research reported in this paper is partially supported under NSERC grant OGPOO44121 and Project B5 of IRIS. The authors wish to thank Craig Boutilier, Andrew Csinger, Mike Horsch, Keiji Kanazawa, Jim Little, Alan Mackworth, Maurice Queyranne, Jack Snoeyink and Ying Zhang for their valuable comments.